\documentclass[runningheads]{llncs}
\usepackage[ruled,vlined]{algorithm2e}
\usepackage[T1]{fontenc}
\usepackage{hyperref}
\usepackage[numbers]{natbib}
\usepackage[table]{xcolor}
\usepackage{multirow, arydshln}
\usepackage{booktabs}
\usepackage{mathrsfs}

\usepackage{amsmath}
\usepackage{graphicx}
\usepackage[nolist]{acronym}
\begin{acronym}[nolist]
    \acro{SRL}{Skeleton Recall Loss}
    \acro{2D}{2 Dimensional}
    \acro{clDice}{centerlineDice}
    \acro{Dice}{Dice loss}
    \acro{DSC}{Dice Similarity Coefficient}
    \acro{JSI}{Jaccard Similarity Index}
    \acro{IoU}{Intersection over Union}
    \acro{95HD}{95th Percentile Hausdorff Distance}
    \acro{ASSD}{Average Symmetric Surface Distance}
    \acro{CCE}{Categorical Cross Entropy loss}
    \acro{FNR}{False Negative Rate}
    \acro{FPR}{False Positive Rate}
    \acro{TP}{True Positive}
    \acro{TN}{True Negative}
    \acro{FP}{False Positive}
    \acro{FN}{False Negative}
    \acro{nnUNet}{no new U-Net}
    \acro{TS}{Tubed Skeletonization}
\end{acronym}

\begin{document}
\title{Does the Skeleton-Recall Loss Really Work?}
%
%
\author{Devansh Arora \thanks{Devansh Arora and Nitin Kumar contributed equally to this work.}
\inst{1} \and
Nitin Kumar \inst{2} \and
Sukrit Gupta \inst{3}}

\authorrunning{Arora and Kumar et al.}
%
\institute{Department of Electrical Engineering, Indian Institute of Technology Ropar, India \and
Department of Computer Science and Engineering, Indian Institute of Technology Ropar, India \and
Biomedical Engineering; and Artificial Intelligence and Data Engineering, Indian Institute of Technology Ropar, India} 

\maketitle              
%

\begin{abstract}
Image segmentation is an important and widely performed task in computer vision. Accomplishing effective image segmentation in diverse settings often requires custom model architectures and loss functions. A set of models that specialize in segmenting thin tubular structures are topology preservation-based loss functions. These models often utilize a pixel skeletonization process claimed to generate more precise segmentation masks of thin tubes and better capture the structures that other models often miss. One such model, \ac{SRL} proposed by Kirchhoff et al.~\cite {kirchhoff2024srl}, was stated to produce state-of-the-art results on benchmark tubular datasets. 
In this work, we performed a theoretical analysis of the gradients for the SRL loss. Upon comparing the performance of the proposed method on some of the tubular datasets (used in the original work, along with some additional datasets), we found that the performance of SRL-based segmentation models is not exceedingly better than traditional baseline models. By providing both a theoretical explanation and empirical evidence, this work critically evaluates the limitations of topology-based loss functions, offering valuable insights for researchers aiming to develop more effective segmentation models for complex tubular structures.

\end{abstract}

\section{Introduction to the problem statement}
Image segmentation models help identify and accentuate structures of interest within an image. These models need to be versatile since they have to accurately delineate objects of diverse shapes, sizes and texture. Thin, tubular and curvilinear structures are particularly challenging for these segmentation models due to a scarce number of pixels corresponding to the region of interest. Common examples of such tasks include: roads in satellite imagery, blood vessels, capillaries \& canals in medical images and histology images. Multiple modifications~\cite{shit2021cldice,cheng2021joint,menten2023skeletonization} have been proposed to efficiently segment thin tubular structures.

One category of such methods is specialized topology preservation-based loss functions that utilize a skeletonization process whereby they reduce a shape in a binary image to its one-pixel-wide connected center-line that preserves its overall topology. These processes are meant to reduce the foreground to a thin network of pixels, which is representative of the topology of the original foreground. Different segmentation methodologies utilize different types of skeletonization processes~\cite{jin20173dskeletonization,abu2013skeletonization,latecki2007skeletonization}, often followed by the employment of skeletons thus formed in a specialized loss function. One such proposed method was the Skeleton Recall Loss~\cite{kirchhoff2024srl}, specialized for the segmentation of thin tubular and curvilinear structures. 

In this work, we investigate the use of skeletonization for topology preservation by examining the \ac{SRL} approach~\cite{kirchhoff2024srl}, which reported state-of-the-art performance for the segmentation of thin tubular structures through the incorporation of skeleton information. We train nnU-Net~\cite{isensee2018nnu} based models on thin tubular structure segmentation tasks, both with and without the \ac{SRL} loss, to systematically assess its impact.

To gain further insight, we conduct a theoretical analysis of the gradients to better understand the optimization behavior of the \ac{SRL} loss in comparison to standard loss functions. In addition, we perform an extensive empirical evaluation across multiple datasets and metrics. Our results indicate that the performance gains offered by \ac{SRL} are limited and dataset-dependent. Specifically, for tubular datasets, out of fifteen evaluated scores (five metrics across three datasets), \ac{SRL} demonstrates statistically significant improvements on three scores, exhibits statistically significant degradation on three scores, and shows marginal changes on six scores. For non-tubular datasets, \ac{SRL} achieves statistically significant improvement on one score, while performance degradation is observed on seven scores, with marginal changes on one score.

\section{Formulation and theoretical analysis of the Skeleton Recall Loss} \label{sec:formulation}
\subsection{Formulation of the skeleton recall loss}
In the paper~\cite{kirchhoff2024srl}, a new segmentation mask image transformation technique is introduced as \ac{TS} (Algorithm \ref{alg:TS}). The procedure first involves forming a skeleton of the ground truth mask using any of the standard skeletonization techniques~\cite{abu2013skeletonization,jin20173dskeletonization,latecki2007skeletonization}. Then, the skeleton is dilated and multiplied with the original ground truth mask (for preserving the class of the pixel), to obtain the tubed skeleton of the ground truth mask.

\begin{figure}[h]
\begin{center}
\includegraphics[width=\textwidth]{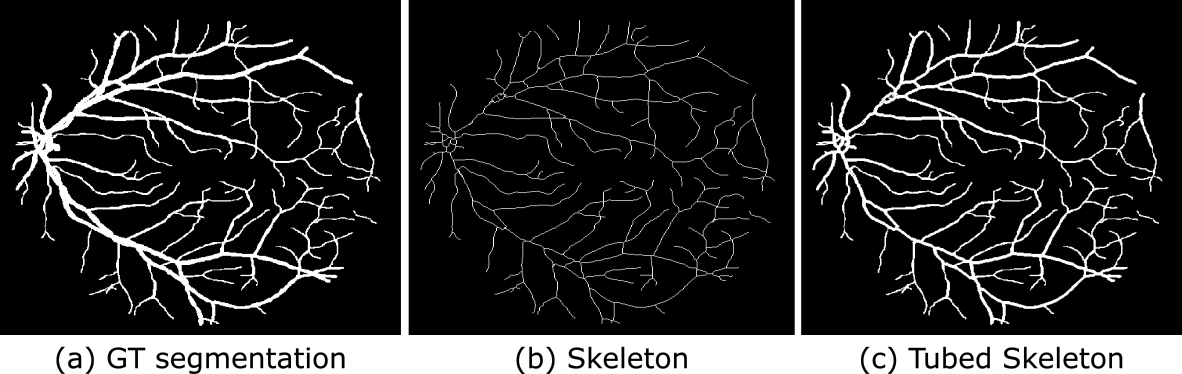}
\end{center}
\caption{Visual comparison of (b) normal skeleton formed using standard skeletonization algorithm and (c) the proposed tubed skeleton(formed by dilating (b) normal skeleton and then multiplying with (a) ground truth) used for Skeleton Recall Loss for (a) ground truth segmentation, originating from the DRIVE dataset~\cite{staal2004ridge}.}
\label{fig: Skeleton}
\end{figure}

The difference between the original mask, a regular skeleton and a tubed skeleton is shown in Figure \ref{fig: Skeleton}. This tubed skeleton is employed as the transformed ground truth to calculate the \ac{SRL} with the following function:
\begin{equation} \label{eqn:SRL}
\mathscr{L}_{SRL} = - \frac{1}{|K|} \sum_{k \in K} \frac{\sum_{i \in \Omega} s_\theta^{ik}.y^{ik}}{\sum_{i \in \Omega} y^{ik}} 
\end{equation}
where $\Omega$ is the set of pixels in segmentation mask, $K$ is the set of all classes present in the ground truth mask,  $s_\theta^{ik}$ denotes the $i^{th}$ pixel of predicted mask for the $k^{th}$ class, and $\theta$ are the parameters that affect $s^{ik}$, and $y^{ik}$ denotes the $i^{th}$ pixel of Tubed Skelton formed using the ground truth mask for the $k^{th}$ class. This loss function is used along with the standard loss functions while training the model. The net loss is then defined as
\begin{equation}
\label{eqn: loss}
\mathscr{L} = \mathscr{L}_{generic} + \alpha \cdot \mathscr{L}_{SRL}
\end{equation}
where $\alpha$ is a hyperparameter that can be tuned to alter the contribution of the specialized loss function towards the model loss. 

\subsection{Gradient analysis of the skeletal recall loss}
To understand the effect of \ac{SRL} on the model training, we propose to investigate the flow of gradients through the neural network due to \ac{SRL} during back propagation. Understanding the gradients would be beneficial for interpreting their effects in the modification of parameters. To investigate the gradient flowing through the network during back propagation, let us consider the gradient of each loss term with respect to the pixels in the predicted mask. From equation \ref{eqn: loss}, the gradient of loss with respect to the $j^{th}$ pixel of predicted mask for the $k^{th}$ class, $s_\theta^{jk}$, which is in turn dependent on parameter $\theta$ is given by:



\begin{equation*}
    \displaystyle \frac{\partial \mathscr{L}} {\partial s_\theta^{jk}} = \frac{\partial \mathscr{L}_{generic}} {\partial s_\theta^{jk}}  + \frac{\partial \mathscr{L}_{SRL}} {\partial s_\theta^{jk}}
\end{equation*}


From equation \ref{eqn:SRL}, the gradient for the \(\mathscr{L}_{SRL}\) term w.r.t. to any predicted pixel $s_\theta^{jk}$ for \ac{SRL} is given by:
\begin{equation*} \label{eqn:SRL_grad}
    \frac{\partial \mathscr{L}_{SRL}}{\partial s_\theta^{jk}} = - \frac{1}{|K|} \frac{y^{jk}}{\sum_{i \in \Omega} y^{ik}}
\end{equation*}

 


Since the tubed skeleton masks $y^{jk}$, that are in turn derived from the ground truth masks, do not change over the epochs, the gradient value, which backpropagates in the neural network, is independent of predicted pixel values $s_\theta^{jk}$. Now, each pixel in the tubed skeleton mask, $y^{jk}$, can either belong to the foreground or to the background. So it can assume only two values: 0 and 1. Hence, the gradient of \ac{SRL} boils down to a simple function:


\begin{equation*}
\frac{\partial \mathscr{L}_{SRL}}{\partial s_\theta^{jk}} =
\begin{cases}
-\frac{1}{\sum_{i \in \Omega} y^{ik}} & \text{if } y^{jk}=1; \\
0 & \text{if } y^{jk} = 0
\end{cases}
\end{equation*}

So, irrespective of the segmentation prediction, \ac{SRL} loss backpropagates a constant value through $j^{th}$ pixel, if its corresponding pixel in the tubed skeleton mask is part of the skeletonized region (foreground), else not. The loss does not include information about the current state of model weights or predictions, which is the root cause of inefficiency of \ac{SRL}.

We further note that each predicted pixel can only belong to one of the four categories: \acf{TP}, \acf{TN}, \acf{FP}, or \acf{FN}. Note that here we refer to a pixel as belonging to a certain category based on its comparison with the original ground truth mask, and not the skeletonized mask. We present an analysis of the gradient of \ac{SRL} for each category of predicted pixel.
\begin{enumerate}
\item \textit{True positive (TP)}:
In case the predicted pixel was a \ac{TP}, we can subdivide the discussion into two further cases. The skeleton is a thinner version of the ground mask, and therefore, a pixel aligning with the positives in the original ground truth mask may or may not align with the positives of the skeletonized mask. Now two cases are possible: 
\begin{itemize}
\item \textbf{The pixel aligns with the positives in the skeletonized mask.} In that case $y^{jk} = 1$. Therefore, a constant gradient keeps flowing throughout the training process, which eventually keeps pushing the parameters $\theta$ for this pixel further away from the already achieved optimal values, leading to poor performance.
\item \textbf{The pixel does not align with the positives in the skeletonized mask.} In this case, $y^{jk} = 0$. Therefore, the parameters $\theta$ affecting this pixel value would only be trained by the generic loss functions and would not experience any alteration due to indulgence of \ac{SRL}. 
\end{itemize}

\item \textit{True negative (TN)}: 
In case the predicted pixel was a \ac{TN}, it would always align with the negative pixels of the skeletonized masks. Therefore, $y^{jk} = 0$ always, implying that the gradient flowing to the parameters through this pixel is unaffected by \ac{SRL}.

\item \textit{False positive (FP)}:
If the predicted pixel is a \ac{FP}, then the pixel aligns with the negative region of the original ground truth mask as well as the skeletonized mask. Similar to case 2, $y^{jk} = 0$ here, and hence the gradient flowing due to \ac{SRL} does not affect the parameters of this pixel.

\item \textit{False negative (FN)}: 
Similar to case 1, if a predicted pixel turns out to be a \ac{FN}, its effect can be studied as two separate cases:
\begin{itemize}
\item \textbf{The pixel aligns with the positives of the skeletonized mask.} In that case, $y^{jk} = 1$, and hence a constant gradient would keep affecting the parameters associated with the chosen pixel. So, even when the generic losses try to alter the parameters in the right direction, \ac{SRL} keeps pushing the gradient in some absurd direction, hence reducing the efficiency of the generic loss functions. 
\item \textbf{The pixel does not align with the positives of the skeletonized mask.} Here, $y^{jk} = 0$, which implies that the \ac{SRL} does not affect the overall gradient of the loss function.
\end{itemize}
\end{enumerate}

Note that predicting a pixel to be a positive either does not affect the \ac{SRL} in case of a \ac{FP}, or it reduces the value of the overall loss heavily in case of a \ac{TP} due to the combined effects of multiple loss functions. Therefore, \ac{SRL} does not penalize for a \ac{FP} but rewards for a \ac{TP}. These incentives push the model to predict more positives. Also, in case of a positive prediction, the gradient of \ac{SRL} affects the parameters only in one of the three possible cases (2 parts of case 1 and case 3), implying that \ac{SRL} does not take care of the \ac{FP} predictions. These factors lead to a significant increase in the \ac{FPR}.

For each parameter $\theta$, the update rule is
\[
\theta_{n} = \theta_{n-1} - \eta \, \frac{\partial \mathscr{L}}{\partial \theta}.
\]

Using the chain rule,
\[
\frac{\partial \mathscr{L}}{\partial \theta}
=
\sum_{j}\sum_{k}
\frac{\partial \mathscr{L}}{\partial s^{jk}_{\theta}}
\;\frac{\partial s^{jk}_{\theta}}{\partial \theta}.
\]

We now study the effect of \ac{SRL} on these parameter updates. Under SRL, whenever $y^{jk} = 0$, the derivative $\frac{\partial \mathscr{L}}{\partial s^{jk}_{\theta}}$ becomes zero; these locations therefore do not contribute to the gradient. When $y^{jk}=1$, the derivative takes a constant negative value, say $-C$. Thus,
\[
\frac{\partial \mathscr{L}_{\mathrm{SRL}}}{\partial \theta}
=
\sum_{j}\sum_{k : y^{jk}=1} (-C)\,\frac{\partial s^{jk}_{\theta}}{\partial \theta}
= -C \sum_{j}\sum_{k} y^{jk}\,\frac{\partial s^{jk}_{\theta}}{\partial \theta}.
\]

Hence, the parameter update becomes
\[
\theta_{n}
= \theta_{n-1} - \eta \frac{\partial \mathscr{L}_{\mathrm{SRL}}}{\partial \theta}
= \theta_{n-1} + \eta C \sum_{j}\sum_{k} y^{jk}\,\frac{\partial s^{jk}_{\theta}}{\partial \theta}.
\]

It is important to emphasize that the network outputs logits that are transformed by a sigmoid or softmax function. Thus, the gradient term changes only due to the derivative of the model outputs with respect to its parameters. Since it depends solely on the logits and not on the discrepancy between prediction and ground truth, SRL does not provide an error-proportional learning signal. Consequently, parameter updates are governed primarily by the sensitivity of the logits with respect to the parameters rather than by a measure of prediction error. This structural property of SRL limits the amount of supervised information transmitted through the gradient and may hinder the model’s ability to improve segmentation quality.

\subsection{Mask Transformation}
Another technical detail that may contribute to the poor performance of the SRL loss function is the accompanying mask transformation.

\begin{algorithm}
    \SetAlgoNlRelativeSize{-1} 
    \KwData{$Y$ is hard class ground truth mask of K classes, such that \(Y_{i,j(,k)} \in [0,K]\)}
    \KwResult{Transformed Mask $Y_{mc-skel}$ of \(K\) classes}
    
        $Y_{bin} \gets Y > 0$\;
        
        $Y_{skel} \gets \textbf{skeletonize}(Y_{bin})$\;

        $Y_{skel} \gets \textbf{dilate}(Y_{skel})$\;

        $Y_{mc-skel} \gets Y_{skel} \times Y$\;
        
    \Return $Y_{mc-skel}$\;
    \caption{Tubed Skeletonization}
    \label{alg:TS}
\end{algorithm}

After the process outlined in Algorithm \ref{alg:TS}, the transformed mask is used for computing \ac{SRL}. The algorithm modifies a simple skeleton to insert additional pixel values into the skeletonized mask. However, after multiplication of the dilated skeleton with the original ground truth, any \ac{FP} introduced during the transformation would be removed as a result of multiplication with the background (which has all pixel values 0). Therefore, the resulting mask becomes overly similar to the ground truth mask, with the only remaining differences being the pixel information removed from thick regions, as shown in Figure \ref{fig: Skeleton}. So, in effect, the transformation does not seem to bring about any significant alterations compared to the original ground truth. The results in Table \ref{tab:mask_results} show that introducing the transformation does not lead to considerable growth in metric values.

\begin{table*}[!ht]
    \centering
    \caption{Test metrics of \ac{SRL} models with/without the mask transformation(\ac{TS}).}
    \begin{tabular}{c : c c c c c c} 
        \toprule
        Datasets & Method & {DSC} \(\uparrow\) & {clDice} \(\uparrow\) & {JSI} \(\uparrow\) & {FNR} \(\downarrow\) & {FPR} \(\downarrow\) \\
        \midrule

        \multirow{2}{*}{DRIVE~\cite{staal2004ridge}}
            & w/o \ac{TS} & $ \mathbf{84.02_{\pm 0.07} }$ & $ 87.81_{\pm 0.08} $ & $ \mathbf{72.48_{\pm 0.11} }$ & $ \mathbf{18.31_{\pm 0.25} }$ & $ 1.73_{\pm 0.04} $ \\ 
            
            & With \ac{TS} & $84.01_{\pm 0.07}$ & $ \mathbf{87.82_{\pm 0.15}} $ & $72.46_{\pm 0.10} $ & $ 18.46_{\pm 0.38} $ & $ \mathbf{1.70_{\pm 0.06} }$ \\

        \midrule

        \multirow{2}{*}{Cracks~\cite{tomaszkiewicz2023cracks}} 
            & w/o \ac{TS} & $76.86_{\pm 0.43} $ & $ 86.05_{\pm 0.56} $ & $ 65.93_{\pm 0.43} $ & $ 17.33_{\pm 0.50} $ & $\mathbf{0.31_{\pm 0.01} }$ \\
            
            & With \ac{TS} & $\mathbf{76.94_{\pm 0.42} }$ & $ \mathbf{86.21_{\pm 0.37} }$ & $\mathbf{ 65.98_{\pm 0.39} }$ & $\mathbf{ 17.09_{\pm 0.50} }$ & $\mathbf{0.31_{\pm 0.01}} $ \\ 

        \bottomrule
    \end{tabular}
    \label{tab:mask_results}
\end{table*}

\section{Results}
\subsection{Model architecture and test metrics}
To replicate the experiments, we employed the nnUNet architecture. Two separate models were trained for each dataset, with each model trained on five random seeds for each dataset. The first vanilla model was trained using standard \ac{Dice} and \ac{CCE} loss functions. The second model was trained on the \(\mathscr{L}_{SRL}\), which combined the \ac{SRL} along with the standard loss functions. The models were trained on NVIDIA RTX A6000 GPU with 50GB of VRAM. The code for this implementation was directly taken from \href{https://github.com/MIC-DKFZ/Skeleton-Recall}{github.com/MIC-DKFZ/Skeleton-Recall}. We included multiple metrics covering aspects of overlap and topology preservation to comprehensively evaluate the performance of the different segmentation models. Among these, \ac{clDice}~\cite{shit2021cldice}, which evaluates the center line of the structures while simultaneously considering overlap, has proven to be a particularly valuable metric for assessing topology preservation. For overlap-based evaluation, we utilized the widely used \ac{DSC} and \ac{IoU} metrics. We used \ac{FNR} to gauge the portions of ground truth that are missed by the model, and hence it gives information about the maintenance of connectivity. The \ac{FPR} was used to ensure the model wasn’t over-predicting positives merely to achieve a low \ac{FNR}.


\begin{table}[!ht]
    \centering
    \caption{\textbf{Dataset Summary} Characteristics of the datasets used for training and evaluation, covering multiple 2D segmentation tasks ranging from binary to multi-class segmentation. Tr and Ts refer to training and test image sets, respectively. \# denotes “Number of”.}
    \begin{tabular}{c : c c c} 
        \toprule
        Dataset & \shortstack{Image \\ Dims} & \shortstack{\# Classes} & \shortstack{\# Images \\ (Tr + Ts)} \\
        
        \midrule
        DRIVE ~\cite{staal2004ridge} & \(512 \times 512\) & 2 & 80 + 20 \\
        
        \midrule
        Roads~\cite{mnih2013roads} & \(512 \times 512\) & 2 & 804 + 13 \\
        
        \midrule
        Cracks ~\cite{tomaszkiewicz2023cracks} & \(224 \times 224\) & 2 & 572 + 143 \\
        
        \midrule
        BoMBR ~\cite{raina2024bombr} & \(512 \times 512\) & 4 & 201 + 50 \\
        
        \midrule
        Drone\footnotemark[1] & \(512 \times 512\) & 5 & 320 + 80 \\
        
        \midrule
        ACDC~\cite{bernard2018acdc} & \( 10 \times 256 \times 216 \) & 3 & 200 + 100 \\
        
        \bottomrule
    \end{tabular}
    \label{tab:Dataset Descriptions}
\end{table}

To validate the effectiveness of \ac{SRL} in segmentation, a wide variety of datasets were utilized by the authors of the SRL paper~\cite{kirchhoff2024srl}. These include natural segmentation tasks like Roads~\cite{mnih2013roads} and Cracks~\cite{tomaszkiewicz2023cracks}, along with medical datasets which include 2D datasets like DRIVE~\cite{staal2004ridge} and 3D datasets like ACDC~\cite{bernard2018acdc}. While the original DRIVE release consists of 40 color fundus photographs (20 training and 20 test images), we utilized the publicly available pre-augmented distribution~\cite{kaggle_drive_retina}, which expands the training set to 80 images via offline data augmentation while retaining the standard 20 test images for final evaluation. 

\subsection{Experiments on tubular datasets}
We tried to reproduce the results of \ac{SRL} on three of the datasets mentioned in the paper~\cite{kirchhoff2024srl}, namely DRIVE~\cite{staal2004ridge}, Cracks~\cite{tomaszkiewicz2023cracks}, and Roads~\cite{mnih2013roads}. DRIVE is a medical dataset focused on segmenting blood vessels from retinal fundus images. Cracks, on the other hand, is a dataset for the segmentation of thin cracks in concrete. The Roads dataset presents a road segmentation task from aerial images. All the datasets are 2D and involve binary classification. Appropriate train-test splits were made for each dataset. The data pre-processing was handled by \ac{nnUNet}~\cite{isensee2018nnu} itself.

The metrics in Table~\ref{tab:Results_tubular} show a mixed comparison between the Vanilla model (trained with Dice and \ac{CCE}) and the \ac{SRL}-based model. Overall, the results indicate that \ac{SRL} does not lead to a consistent or substantial improvement on the tubular datasets. In the case of the DRIVE dataset, its use is associated with a decline in performance.

\begin{figure}[!ht]
\begin{center}
\includegraphics[width=\textwidth]{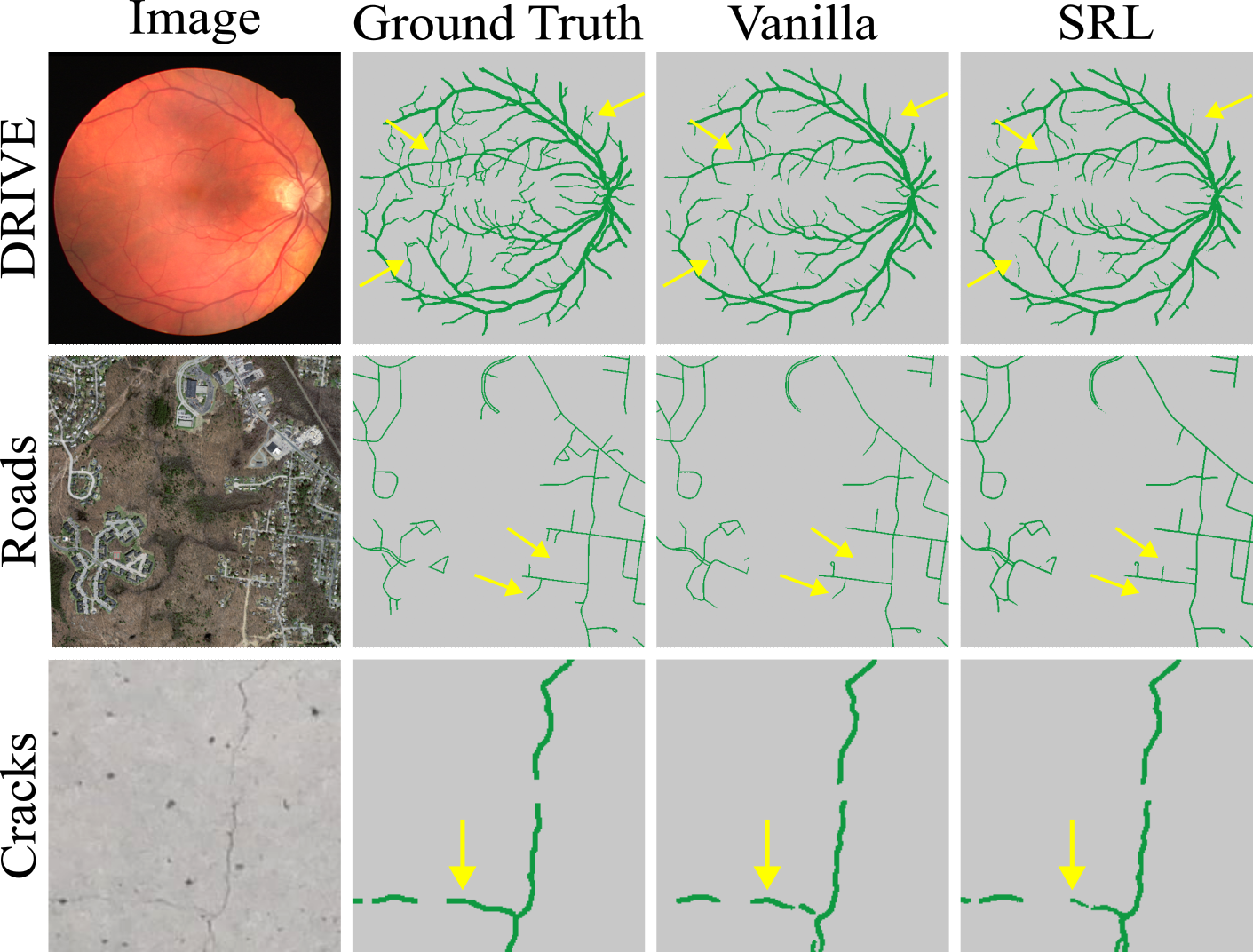}
\end{center}
\caption{Results of the proposed method over 2 of the datasets. \ac{nnUNet} with conventional segmentation losses performs adequately in mapping the basic structure of the object of interest, often outperforming the \ac{SRL} method, as demonstrated in examples from (a) DRIVE and (b) Cracks datasets.}
\label{fig:results_tubular}
\end{figure}

The qualitative outputs in Figure~\ref{fig:results_tubular} further support these observations. Both Vanilla and \ac{SRL} models perform comparably well on thicker tubular structures; however, several thin structures captured by the Vanilla models are missed by the \ac{SRL} variants. 

The statistical significance analysis in Table~\ref{tab:significance_stats} reinforces these trends: the \ac{SRL} model shows significant improvement in only 3 out of the 15 evaluated metrics (5 metrics across 3 datasets). At the same time, it significantly worsens the \ac{FPR} across all datasets. Although the \ac{SRL} model achieves a noteworthy reduction in \ac{FNR}, this gain is counteracted by a corresponding significant rise in \ac{FPR}. This behavior suggests a tendency toward over-segmentation when \ac{SRL} is used.

Taken together, these findings suggest that, within the evaluated settings, \ac{SRL}-based models do not consistently outperform models trained with more traditional loss functions.

\begin{table*}[!ht]
    \centering
    \caption{\textbf{Test set metrics} of nnUNet models trained on tubular datasets with varying loss functions. An asterisk ($\mathbf{*}$) indicates statistical significance of the marked metric with respect to the other metric at $p < 0.05$, based on t-tests.}
    \begin{tabular}{c : c c c c c c} 
        \toprule
        Datasets & Method & {DSC} \(\uparrow\) & {clDice} \(\uparrow\) & {JSI} \(\uparrow\) & {FNR} \(\downarrow\) & {FPR} \(\downarrow\) \\
        \midrule

        \multirow{2}{*}{DRIVE~\cite{staal2004ridge}}
            & UNet & $ \mathbf{84.05_{\pm 0.10}}$ & $ \mathbf{87.91_{\pm 0.13}}$ & $ \mathbf{72.53_{\pm 0.14}} $ & $ 18.92_{\pm 0.32}$ & $\mathbf{1.06_{\pm 0.03}}^*$ \\

            & SRL & $ 84.01_{\pm 0.07} $ & $ 87.82_{\pm 0.15} $ & $72.46_{\pm 0.10} $ & $ \mathbf{18.46_{\pm 0.38}} $ & $ 1.70 _{\pm 0.06} $ \\

        \midrule

        \multirow{2}{*}{Roads~\cite{mnih2013roads}} 
            & UNet & $ 75.56_{\pm 0.09}  $ & $ 84.23_{\pm 0.12} $ & $ 61.03_{\pm 0.11}  $ & $ 28.37_{\pm 0.36}  $ & $\mathbf{ 1.35_{\pm 0.04} }^*$ \\
            
            & SRL & $ \mathbf{75.84_{\pm 0.38} }$ & $\mathbf{ 85.28_{\pm 0.17} }^*$ & $ \mathbf{61.37_{\pm 0.48}  }$ & $ \mathbf{23.89_{\pm 1.30}}^* $ & $ 1.83_{\pm 0.23}  $ \\

        \midrule

        \multirow{2}{*}{Cracks~\cite{tomaszkiewicz2023cracks}} 
            & UNet & $76.64_{\pm 0.25} $ & $ 85.67_{\pm 0.53} $ & $ 65.67_{\pm 0.29} $ & $ 20.21_{\pm 0.68} $ & $\mathbf{0.25_{\pm 0.01} }^*$ \\
            
            & SRL & $\mathbf{76.94_{\pm 0.42} }$ & $ \mathbf{86.21_{\pm 0.37} }$ & $\mathbf{ 65.98_{\pm 0.39} }$ & $\mathbf{ 17.09_{\pm 0.50} }^*$ & $0.31_{\pm 0.01} $ \\

        \bottomrule
    \end{tabular}
    \label{tab:Results_tubular}
\end{table*}
The visual results also offer important insight into model behavior. Although \ac{SRL} was designed to enhance the segmentation of thin tubular structures, the qualitative outputs suggest that this improvement does not consistently materialize. As illustrated in Figure~\ref{fig:results_tubular}, several thin structures captured by the Vanilla models are not recovered by \ac{SRL}. Additionally, \ac{SRL} occasionally produces extra tubular-like regions that are not present in the ground truth, an observation aligned with its higher \ac{FPR}, particularly noticeable in the Roads dataset. These findings highlight that, in the considered settings, \ac{SRL} may not generalize as effectively to fine-grained tubular patterns as initially expected.
\subsection{Experiments on non-tubular datasets}
While \ac{SRL} claimed to improve the segmentation of tubular structures, we tried to analyze its effect on non-tubular datasets as well. This was attempted to understand the generalizability of the methodology, and in what areas it brings significant differences. We performed experiments on non-tubular datasets, like ACDC~\cite{bernard2018acdc}, BoMBR~\cite{raina2024bombr}, and Drone\footnotemark[1]. The Automatic Cardiac Diagnosis Challenge (ACDC)~\cite{bernard2018acdc}, which is a popular non-tubular 3D dataset for research in fields related to cardiac diagnosis and includes 300 multi-equipment CMRI recordings with the task of segmenting the Left Ventricle, Right Ventricle, and Myocardium dataset. BoMBR~\cite{raina2024bombr}, a 2D dataset, consists of reticulin-stained histology images for segmenting different components of a bone marrow biopsy report. Drone\footnotemark[1] presents segmentation of objects in aerial drone images. All datasets involve structures of irregular shapes and sizes, testing the effect of \ac{SRL} in alternate domains. 

In the case of the non-tubular datasets (Table~\ref{tab:Results_nontubular}), the \ac{SRL}-based models tend to underperform relative to the Vanilla models across all datasets. These results suggest that for datasets containing a wide range of complex mask geometries, such as Drone, the topology-based loss does not confer a clear advantage. The statistical tests further support this observation, indicating that the \ac{SRL} model performs comparably to, or worse than, the Vanilla model in most settings. Specifically, 7 out of the 15 evaluated metrics (5 metrics across 3 datasets) show a significant advantage for the Vanilla models.

\begin{figure}[!ht]
\begin{center}
\includegraphics[width=\textwidth]
{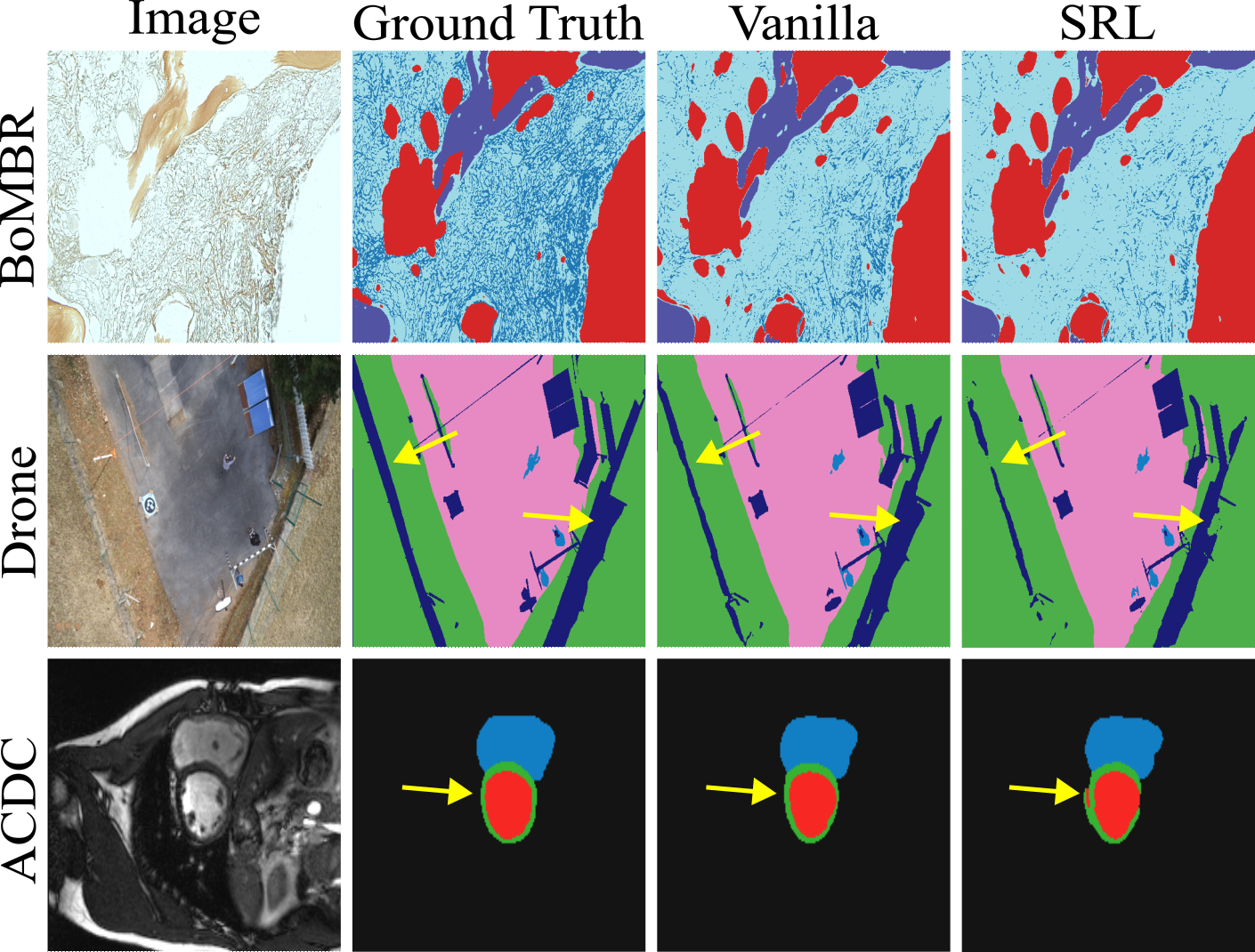}
\end{center}
\caption{Results of proposed method over other datasets of alternate domains, namely- (a) BoMBR~\cite{raina2024bombr}, and (c) Drone{\protect\footnotemark[1]}. In the BoMBR dataset, \ac{SRL} model lacks significantly in the prediction of reticulin fibres. On a similar note, it also fails to predict basic shapes accurately when introduced with noisy real-world images from the Drone dataset or with 3D medical images.}
\label{fig:results_nontubular}
\end{figure}

The qualitative examples in Figure~\ref{fig:results_nontubular} provide additional insight. For the BoMBR dataset, the \ac{SRL} model struggles to segment the reticulin fibers, which exhibit thin, tubular patterns. In the Drone dataset, the use of \ac{SRL} appears to introduce additional mispredictions, potentially due to its sensitivity to shape complexity. Similarly, in the ACDC dataset, the \ac{SRL} model has difficulty delineating relatively simple blob-like structures, whereas the Vanilla model captures these boundaries more accurately.

Overall, these results indicate that \ac{SRL} does not consistently enhance performance on non-tubular datasets and may, in some cases, hinder accurate segmentation. However, it is to be noted that \ac{SRL} was not intended to work on non-tubular datasets, and these results have just been presented for analysis purpose.



\begin{table*}[!ht]
    \centering
    \caption{\textbf{Test set metrics} of nnUNet models trained on non-tubular datasets with varying loss functions. An asterisk ($\mathbf{*}$) indicates statistical significance of the marked metric with respect to the other metric at $p < 0.05$, based on t-tests.}
    \begin{tabular}{c : c c c c c c} 
        \toprule
        Datasets & Method & {DSC} \(\uparrow\) & {clDice} \(\uparrow\) & {JSI} \(\uparrow\) & {FNR} \(\downarrow\) & {FPR} \(\downarrow\) \\
        \midrule

        \multirow{2}{*}{BoMBR~\cite{raina2024bombr}}
            & UNet & $\mathbf{69.15_{\pm 0.51} }$ & $\mathbf{66.67_{\pm 0.41} }$ & $\mathbf{62.16_{\pm 0.43} }$ & $\mathbf{30.56_{\pm 0.44}} $ & $7.16_{\pm 0.17} $ \\
            
            & SRL & $69.04_{\pm 1.16} $ & $ 66.51_{\pm 1.17} $ & $62.11_{\pm 1.01} $ & $ 30.82_{\pm 0.66} $ & $\mathbf{ 7.01_{\pm 0.33} }$ \\

        \midrule

        \multirow{2}{*}{Drone\footnotemark[1]} 
            & UNet & $\mathbf{87.40_{\pm 0.81}}^*$ & $\mathbf{ 83.18_{\pm 0.97}}^*$ & $ \mathbf{80.22_{\pm 0.91}}^*$ & $ \mathbf{11.87_{\pm 0.25}}$ & $\mathbf{2.01_{\pm 0.04}}^*$ \\

            & SRL & $85.42_{\pm 1.36} $ & $ 81.21_{\pm 1.25} $ & $ 78.19_{\pm 1.25} $ & $ 12.06_{\pm 0.45} $ & $2.38_{\pm 0.28}$ \\
            
        \midrule

        \multirow{2}{*}{ACDC~\cite{bernard2018acdc}}
            & UNet & $\mathbf{91.88_{\pm 0.06}}^*$ & $\mathbf{94.64_{\pm 0.05}}$ & $ \mathbf{85.35_{\pm 0.10}}^*$ & $\mathbf{7.94_{\pm  0.08}}^*$ & $ 0.09_{\pm 0.00} $ \\
            
            & SRL & $91.77_{\pm 0.08} $ & $ 94.51_{\pm 0.19} $ & $ 85.17_{\pm 0.13} $ & $ 8.40_{\pm 0.15} $ & $\mathbf{ 0.08_{\pm 0.00}}^*$ \\
        
        \bottomrule
    \end{tabular}
    \label{tab:Results_nontubular}
\end{table*}

\setcounter{footnote}{1}
\footnotetext{\url{http://dronedataset.icg.tugraz.at/}}

\subsection{Effectiveness of \ac{SRL} on 3D Tubular Datasets}
Although our analysis is limited to 2D datasets, it is relevant to reflect on the performance trends reported for \ac{SRL} in 3D settings. The original work demonstrates comparatively better results on 3D tubular datasets; however, this does not directly resolve the concerns raised in our gradient analysis. The core mechanism of \ac{SRL}, namely, the constant gradient contribution for foreground pixels and the absence of corrective gradients for background pixels, remains unchanged when moving from 2D to 3D. From a theoretical standpoint, the same gradient dynamics should continue to bias the model towards over-predicting foreground structures irrespective of dimensionality.

In light of this, the improved performance observed in 3D datasets warrants further investigation. It is possible that dataset-specific factors, such as stronger spatial continuity or reduced ambiguity in volumetric shapes, mitigate some of the limitations exposed in 2D. However, such hypotheses require dedicated empirical validation. Our 2D results highlight that the mathematical behavior of \ac{SRL} may not universally translate into better segmentation performance, and a more comprehensive examination on diverse 3D datasets is necessary to isolate whether the reported gains arise from the loss function itself or from characteristics intrinsic to the datasets used.
\section{Constant Gradient: An Error or An Aid?}

While \ac{SRL} has been proposed as a topology-aware loss for tubular structures, several fundamental questions arise regarding its mechanism and applicability:

\begin{enumerate}
    \item \textbf{Given the constant gradient, is it possible that SRL's observed improvements are coincidental in certain cases?} \\[2mm]
    Our analysis shows that the gradient of \ac{SRL} with respect to foreground pixels is a constant, while background pixels receive no suppressive gradient. This structure creates a bias towards over-predicting foreground regions, independent of the actual correctness of the predictions. Therefore, any observed gains in Recall or segmentation performance may arise from dataset-specific characteristics, such as sparsity, object size, or connectivity patterns, rather than a generalizable property of the loss function itself. The same constant gradient mechanism applies in both 2D and 3D, raising the question of whether improvements reported in certain 3D datasets are truly reflective of the loss formulation or are instead influenced by particular dataset properties.

    \item \textbf{Is such a constant gradient an aid for the segmentation of tubular structures?} \\[2mm]
    The constant gradient enforces uniform positive updates for all foreground pixels, but it does not inherently incorporate error-sensitive feedback that reflects the distance from the ground truth. While this may encourage the network to predict connected structures, it is unclear whether this is an aid for accurate tubular segmentation. Could forcing the model to predict more \ac{FP} be beneficial for such segmentation tasks? Other loss formulations, such as Dice or cross-entropy, provide gradient signals proportional to the prediction error and may achieve similar or better performance without the risk of over-prediction. The theoretical behavior of \ac{SRL} thus invites scrutiny regarding whether a constant gradient is an aid, or whether alternative mechanisms could provide more precise guidance.
\end{enumerate}

These questions highlight the need for further empirical and theoretical investigation to fully understand the conditions under which \ac{SRL} is effective, and the extent to which its gradient mechanics influence the observed performance.

\section{Conclusion}
In this paper, we evaluated a skeletonization-based segmentation methodology, \acf{SRL}, which incorporates a mask transformation and a specialized loss function aimed at improving the segmentation of thin tubular structures. Our analysis was conducted on two levels: theoretical and empirical.

The theoretical analysis examined the gradients propagated through the model due to \ac{SRL} and its associated mask transformation. We found that the constant gradient associated with \ac{SRL} encourages a higher number of positive predictions, contributing to an increase in \ac{FPR}. The mask transformation also results in a loss of information that may influence the training dynamics. Table~\ref{tab:mask_results} shows that applying \ac{SRL} without the mask transformation achieves comparable performance on tubular datasets, suggesting that the transformation does not significantly alter the outcomes.

The empirical results in Tables~\ref{tab:Results_tubular} and~\ref{tab:Results_nontubular} are consistent with these theoretical observations. The application of \ac{SRL} with UNet does not consistently improve performance, and in some cases, the segmentation quality is reduced. Evaluations on non-tubular datasets further indicate that the effectiveness of \ac{SRL} may be limited to specific types of data, and its benefits are not necessarily generalizable across diverse datasets.

Together, these analyses provide a comprehensive perspective on the behavior of \ac{SRL}, highlighting both its potential and its limitations in different contexts, and emphasizing the importance of carefully considering dataset characteristics when applying topology-aware loss functions.

\bibliographystyle{splncs04}
\bibliography{main}
\appendix
\section{Statistical Significance Test}
To test the statistical significance of the generated results, we employ a two-sample one-sided t-test. This test takes into consideration all the variance in the test metrics of the baseline as well as the test subject. Hence, it can capture the differences brought about by the proposed methodology. Table \ref{tab:significance_stats} presents the t-values and p-values obtained for the t-test on each metric for every dataset. 

\begin{table*}[ht]
\centering
\caption{\textbf{Statistical significance tests} of metric values compared between SRL and the Vanilla UNet base model.}
\begin{minipage}{0.48\textwidth}
\centering
\begin{tabular}{l : l c c}
    \toprule
    \textbf{Dataset} & \textbf{Metric} & \textbf{t-value} & \textbf{p-value} \\
    \midrule
    \multirow{5}{*}{DRIVE\cite{staal2004ridge}} 
        & DSC    &  -0.730 & 0.756 \\
        & \ac{clDice} &  -0.8252 & 0.783 \\
        & JSI    &  -0.765 & 0.766 \\
        & FNR    &  -1.86 & 0.051 \\
        & FPR    & 3.086 & 0.990 \\
    \midrule
    \multirow{5}{*}{Cracks\cite{tomaszkiewicz2023cracks}} 
        & DSC    & 1.238 & 0.129 \\
        & \ac{clDice} & 1.657 & 0.070 \\
        & JSI    & 1.257 & 0.124 \\
        & FNR    & -7.396 & 0.000 \\
        & FPR    & 9.622 & 1.000 \\
    \midrule
    \multirow{5}{*}{Roads\cite{mnih2013roads}} 
        & DSC    & 1.464 & 0.105\\
        & \ac{clDice} & 10.107 & 0.000\\
        & JSI    & 1.376 & 0.117\\
        & FNR    & -6.62 & 0.001 \\
        & FPR    & 4.157 & 0.994 \\
    \bottomrule
\end{tabular}
\end{minipage}
\hfill
\begin{minipage}{0.48\textwidth}
\centering
\begin{tabular}{l : l c c}
    \toprule
    \textbf{Dataset} & \textbf{Metric} & \textbf{t-value} & \textbf{p-value} \\
    \midrule
    \multirow{5}{*}{BoMBR\cite{raina2024bombr}} 
        & DSC    & -0.167 & 0.562\\
        & \ac{clDice} & -0.253 & 0.595\\
        & JSI    & -0.088 & 0.533\\
        & FNR    & 0.668 & 0.737\\
        & FPR    & -0.821 & 0.222\\
    \midrule
    \multirow{5}{*}{Drone\footnotemark[1]} 
        & DSC    & -2.498 & 0.978\\
        & \ac{clDice} & -2.493 & 0.980\\
        & JSI    & -2.633 & 0.984\\
        & FNR    & 0.748 & 0.759\\
        & FPR    & 2.611 & 0.972\\
    \midrule
    \multirow{5}{*}{ACDC\cite{bernard2018acdc}} 
        & DSC    & -2.192 & 0.969\\
        & \ac{clDice} & -1.384 & 0.885\\
        & JSI    & -2.085 & 0.963\\
        & FNR    & 5.441 & 0.999\\
        & FPR    & -5.758 & 0.000\\
    \bottomrule
\end{tabular}
\end{minipage}
\label{tab:significance_stats}
\end{table*}

These metrics test the statistical significance of \ac{SRL}, considering the Vanilla model as baseline. The metrics where p-value < 0.05 are considered statistically significant. However, if one wishes to capture the substantial improvement in Vanilla compared to \ac{SRL}, it is evident via p-values > 0.95, as determined by the same two-sample t-test. This would be equivalent to looking for p-values < 0.05 in a reversed t-test, where \ac{SRL} is considered the baseline for testing results of the Vanilla model.

\end{document}